\documentclass[10pt,twocolumn,letterpaper]{article}
\usepackage{iccv}
\usepackage{times}
\usepackage{epsfig}
\usepackage{graphicx}
\usepackage{subfigure} 
\usepackage{enumitem}
\usepackage{epstopdf}
\usepackage{enumitem}
\usepackage{amsmath}
\usepackage{amssymb}
\usepackage{amsfonts}
\usepackage{amsthm}
\usepackage{subfigure}
\usepackage{todo}
\usepackage{verbatim}

\newcommand{\circR}{\operatornamewithlimits{circ}}

\newcommand{\br}{\mathbf{r}}
\newcommand{\bx}{\mathbf{x}}
\newcommand{\bw}{\mathbf{w}}
\newcommand{\bR}{\mathbf{R}}

\usepackage[breaklinks=true,bookmarks=false]{hyperref}
\iccvfinalcopy 

\def\iccvPaperID{652}

\ificcvfinal\fi
\begin{document}

\title{An Exploration of Parameter Redundancy in Deep Networks with \\ Circulant Projections}

\author{ 
Yu Cheng$^{1,3}$\thanks{indicates equal contributions.}\,\,  \hspace{0.1in} Felix X. Yu$^{2,4*}$  \hspace{0.1in}Rogerio S. Feris$^1$   \hspace{0.1in}Sanjiv Kumar$^2$  \hspace{0.1in}Alok Choudhary$^3$  \hspace{0.1in}Shih-Fu Chang$^4$\\ \\
$^1$IBM Research \hspace{0.2in} $^2$Google Research \hspace{0.2in} $^3$Northwestern University \hspace{0.2in} $^4$Columbia University
}
\maketitle

\begin{abstract}
We explore the redundancy of parameters in deep neural networks by replacing the conventional linear projection in fully-connected layers with the circulant projection. The circulant structure substantially reduces memory footprint and enables the use of the Fast Fourier Transform to speed up the computation. Considering a fully-connected neural network layer with $d$ input nodes, and $d$ output nodes, this method improves the time complexity from $\mathcal{O}(d^2)$ to $\mathcal{O}(d\log{d})$ and space complexity from $\mathcal{O}(d^2)$ to $\mathcal{O}(d)$. The space savings are particularly important for modern deep convolutional neural network architectures, where fully-connected layers typically contain more than 90\% of the network parameters.
We further show that the gradient computation and optimization of the circulant projections can be performed very efficiently. 
Our experiments on three standard datasets show that the proposed approach achieves this significant gain in storage and efficiency with minimal increase in error rate compared to neural networks with unstructured projections.
\end{abstract}

\section{Introduction}
\label{sec:intro}

Deep neural network-based methods have recently achieved dramatic accuracy improvements in many areas of computer vision, including image classification \cite{krizhevsky2012, zeiler2014,chen2014}, 
object detection \cite{girshick2014,sermanet2014}, face recognition \cite{yaniv2014,sun2014},  and text recognition \cite{bissacco2014,jaderberg2014_text}.

These high-performing methods rely on deep networks containing  millions or even billions of parameters. For example, the work by Krizhevsky \emph{et al.} \cite{krizhevsky2012}
achieved breakthrough results on the 2012 ImageNet challenge using a network containing 60 million parameters with five 
convolutional layers and three fully-connected layers. The top face verification results  on the Labeled Faces in the Wild (LFW) dataset were obtained with networks containing hundreds of  millions of parameters, using a mix of convolutional, locally-connected, and
fully-connected layers \cite{yaniv2014, sun2015}. In architectures that rely only on fully-connected layers, the number of parameters can grow to billions \cite{dean2012}.

As larger neural networks are considered, with more layers and also more nodes in each layer, reducing their storage and computational costs becomes critical to meeting the requirements of practical applications. Current efforts towards this goal focus mostly on the optimization of convolutional layers \cite{jaderberg2014,mathieu2013fast,erhan2014}, which consume the bulk of computational processing in modern convolutional architectures. We instead explore the redundancy of parameters in fully-connected layers, which are often the bottleneck in terms of memory consumption. Our solution relies on a simple and efficient approach based on \emph{circulant projections} to significantly reduce the storage and computational costs of fully-connected neural network layers, while mantaining competitive error rates. Our work brings the following advantages:

\begin{itemize}[leftmargin=*]
\item Fully-connected layers in modern convolutional architectures typically contain over
90\% of the network parameters. Our approach significantly reduces the cost to store
these networks in memory, which is crucial for GPUs or embedded systems with tight memory constraints.
\item The proposed method enables FFT-based computation, which speeds up evaluation of fully connected layers. This is especially useful for neural networks with many fully connected layers, or consisting exclusively of fully connected layers \cite{soltau2014, dean2012}.
\item With much fewer parameters, our method is empirically shown to require less training data.
\end{itemize}

\subsection{Overview of the proposed approach}
A basic computation in a fully-connected neural network layer is
\begin{equation}
h(\mathbf{x}) = \phi (\mathbf{R} \mathbf{x}),
\label{eq:intro}
\end{equation}
where $\mathbf{R} \in \mathbb{R}^{k \times d}$, and $\phi(\cdot)$ is a element-wise nonlinear activation function.
The above operation connects a layer with $d$ nodes, and a layer with $k$ nodes. 
In convolutional neural networks, the fully connected layers are often used before the final softmax output layer, in order to capture global properties of the image. 
The computational complexity and space complexity of this linear projection are $\mathcal{O}(dk)$. 
In practice, $k$ is usually comparable or even larger than $d$. This leads to computation and space complexity at least $\mathcal{O}(d^2)$, creating a bottleneck for many neural network architectures. 

In this work, we propose to impose a \emph{circulant structure} on the projection matrix $\mathbf{R}$ in (\ref{eq:intro})\footnote{A sign flipping operation is applied before the circulant projection matrix. We will present the formal framework in Section \ref{sec:framework}}. 
\begin{eqnarray}
h(\mathbf{x}) = \phi (\mathbf{R} \mathbf{x}), \quad \text{$\mathbf{R}$ is a circulant matrix.}
\end{eqnarray}
This special structure  dramatically reduces the number of parameters. It also allows us to use the Fast Fourier Transform (FFT) to speed up the computation. Considering a neural network layer with $d$ input nodes, and $d$ output nodes, the proposed method reduces the space complexity from $\mathcal{O}(d^2)$ to $\mathcal{O}(d)$, and the  time complexity from $\mathcal{O}(d^2)$ to $\mathcal{O}(d \log d)$. Table \ref{table:methods} compares the time and space complexity of the proposed approach with the conventional method.

Surprisingly, although the circulant matrix is highly structured with a very small number of parameters ($\mathcal{O}(d)$), it captures the global information well, and does not impact the final performance much. We show empirically that our method can provide significant reduction of storage and computational costs while achieving very competitive error rates.

\subsection{Organization}
Our work is organized as follows.  
We propose imposing the circulant structure on the linear projection matrix of fully-connected layers of neural networks to speed up computations and reduce storage costs in Section \ref{sec:cnn}.
We show a method which can efficiently optimize the neural network while keeping the circulant structure in Section \ref{sec:opt}. 
We demonstrate with experiments on visual data that the proposed method can speed up the computation and reduce memory needs while maintaining competitive error rates in Section \ref{sec:exp}. We begin by reviewing related work in the following section.

\begin{table}
\begin{center}
\small
\begin{tabular}{l|l|l|l}
\hline Method & Time   & Space  & Time (Learning)\\ 
\hline   Conventional NN    & $\mathcal{O}(d^2)$ & $\mathcal{O}(d^2)$ & $\mathcal{O}(n t d^2)$\\ 
\hline   Circulant NN & $\mathcal{O}(d\log{d})$ & $\mathcal{O}(d)$  & $\mathcal{O}(n t d\log{d})$\\ 
\hline 
\end{tabular}
\end{center}
\caption{Comparison of the proposed method with neural networks based on unstructured projections. 
We assume a fully-connected layer, and the number of input nodes and number of output nodes are both $d$. $t$ is the number of gradient steps in optimizing the neural network. }
\label{table:methods}
\end{table}

\section{Related Work}
\label{sec:related}

\subsection{Deep Learning}

In the past few years, deep neural network methods have achieved impressive results in many visual recognition tasks \cite{krizhevsky2012,yaniv2014,girshick2014,sermanet2014}. Recent advances on learning these models include the use of drop-out~\cite{dropout2012} to prevent overfitting, more effective non-linear activation functions such as rectified linear units~\cite{relu2011} or max-out~\cite{goodfellow2013maxout}, and richer modeling through Network in Network (NiN)~\cite{chen2014}. 
In particular, training high-dimensional networks with large quantities of training data is key to obtaining good results, but at the same time incurs increased computation and storage costs.

\subsection{Compressing Neural Networks}

The work of Collins and Kohli \cite{collins2014} addresses the problem of memory usage in deep networks by  
applying sparsity-inducing regularizers during training to encourage zero-weight connections in 
the convolutional and fully-connected layers. Memory consumption is reduced only at test time, whereas our method cuts down storage costs at both training and testing times.
Other approaches exploit low-rank matrix factorization \cite{sainath2013, denil2013} to reduce the number of neural network parameters. In contrast, our approach exploits the redundancy in the parametrization of deep architectures by imposing a circulant structure on the projection matrix, reducing its storage to a single column vector, while allowing the use of FFT for faster computation.

Techniques based on {\em knowledge distillation} \cite{hinton2015} aim to compress the knowledge of a 
network with a large set of parameters into a compact and fast-to-execute network model. This can 
be achieved by training a compact model to imitate the soft outputs of a larger model. Romero et 
al \cite{romero2015} further show that the intermediate representations learned by the large model
serve as hints to improve the training process and final performance of the compact model.
In contrast, our work does not require the training of an auxiliary model.

Network in Network \cite{chen2014} has been recently proposed as a tool for richer local patch modeling in convolutional networks, where linear convolutions in each layer are replaced by convolving the input with a {\em micro-network} filter defined, for example, by a multi-layer perceptron. The inception architecture \cite{szegedy2014} extends this work by using these micro-networks as dimensionality reduction modules to remove computational bottlenecks and reduce storage costs. A key differentiating aspect is that we focus on modeling global dependencies and reducing the cost of fully connected layers, which usually contain the large majority of parameters in standard configurations. Therefore, our work is complementary to these methods. Although \cite{chen2014} suggests that fully-connected layers could be replaced by average pooling without hurting performance for general image classification, other works in computer vision \cite{yaniv2014} and speech recognition \cite{soltau2014} highlight the importance of these layers to capture global dependencies and achieve state-of-the-art results.

\subsection{Speeding up Neural Networks}
Several recent methods have been proposed to speed-up the computation of neural networks, with focus on convolutional architectures \cite{jaderberg2014,mathieu2013fast,erhan2014, denton2014exploiting}. Related to our work, Mathieu \etal \cite{mathieu2013fast} use the Fast Fourier Transform to accelerate the computation of convolutional layers,
through the well-known convolution theorem. In contrast, our work is focused on the optimization of fully-connected layers by imposing  circulant structure on the weight matrix to speed up the computation in both training and testing stages. 

In the context of object detection, many techniques such as detector cascades or segmentation-based selective search \cite{sande2011,erhan2014} have been proposed to reduce the number of candidate object locations in which a deep neural network is applied. Our proposed approach is complementary to these techniques. 

Other approaches for speeding up neural networks rely on hardware-specific optimizations. For example, fast neural network implementations have been proposed for GPUs \cite{donahue2013decaf}, CPUs \cite{vanhoucke2011improving}, FPGAs \cite{farabet2010hardware}, and on-chip implementations \cite{prezioso2015training}.

Our method is also related to the recent efforts around``shallow'' neural networks, which show that sometimes  shallow structures can match the performance of deep structures \cite{mcdonnell2015enhanced, mcdonnell2014fast, huang2014kernel, yu2015compact}.

\subsection{Linear Projection with Structured Matrices}
Structured matrices have been used in improving the 
space and computational complexities for different learning paradigms. 
For example, circulant matrices have been used in 
dimensionality reduction \cite{vybiral2011variant}, 
binary embedding \cite{yu2014_cbe} and kernel approximation  \cite{yu2015compact}. It has been shown that circulant structure can be used to save space and computation costs without performance degradation. 
The properties of circulant matrices have also been exploited to avoid expensive rounds of hard negative mining in training of object detectors \cite{henriques2013} and for real-time tracking \cite{henriques2012}.

One could in principle use other structured matrices such as Hadamard matrices along with a sparse random Gaussian matrix to achieve fast projection as was done in the fast Johnson-Lindenstrauss transform \cite{ailon2006approximate, dasgupta2011fast}, but they are slower than the circulant projection and need more space. We note that very recently, this idea has been studied in \cite{yang2014deep}.

\section{Circulant Neural Network Model}
\label{sec:cnn}

In this section, we present the general framework of the circulant neural network model, showing the advantages of this model in achieving more efficient computational processing and storage cost savings. 

\subsection{Framework}
\label{sec:framework}

A circulant matrix $\mathbf{R} \in \mathbb{R}^{d \times d}$ is a matrix defined by a vector $\mathbf{r}  = (r_0, r_1, \cdots,  r_{d-1})$:

\begin{small}
\begin{align}
\mathbf{R} = \circR(\mathbf{r}) :=
\begin{bmatrix}
r_0     & r_{d-1} & \dots  & r_{2} & r_{1}  \\
r_{1} & r_0    & r_{d-1} &         & r_{2}  \\
\vdots  & r_{1}& r_0    & \ddots  & \vdots   \\
r_{d-2}  &        & \ddots & \ddots  & r_{d-1}   \\
r_{d-1}  & r_{d-2} & \dots  & r_{1} & r_{0}
\end{bmatrix}.
\label{eq:cir}
\end{align}
\end{small}

Let $\mathbf{D}$ be a diagonal matrix with each diagonal entry being a Bernoulli  variable ($\pm 1$ with probability 1/2).
For $\mathbf{x} \in \mathbb{R}^d$, its $d$-dimensional output is:
\begin{align}
h(\mathbf{x}) = \phi (\mathbf{R} \mathbf{D} \mathbf{x}), \quad \mathbf{R} = \circR(\mathbf{r}).
\label{eq:def}
\end{align}

The projection with the matrix $\mathbf{D}$ corresponds to a random sign flipping step on the data.

\subsection{Motivation}
The idea of replacing unstructured projection with circulant projection is motivated by using circulant projections in dimensionality reduction \cite{vybiral2011variant}, binary embedding \cite{yu2014_cbe}, and kernel approximation \cite{yu2015compact}. From the efficiency point of view, this structure creates great advantages in both space and computation (detailed in Section \ref{subsec:space_time}), and  enables efficient optimization procedures (Section \ref{sec:opt}). 

It is shown in previous works \cite{hinrichs2011johnson, vybiral2011variant, yu2014_cbe, yu2015compact} 
that when the parameters of the circulant projection matrix are generated \emph{iid.} from the standard normal distribution, the circulant projection (with the random sign flipping matrix) mimics an unstructured randomized projection. 
In other words, randomized circulant projections can also be used in different frameworks to preserve pairwise $\ell_2$ distance \cite{hinrichs2011johnson, vybiral2011variant}, angle \cite{yu2014_cbe}, and shift-invariant kernels \cite{yu2015compact}. 
It is then reasonable to conjecture that randomized circulant projections can also achieve good performance in neural networks (compared to using unstructured randomized matrices). This is indeed true, as we will demonstrate empirically. And similar to binary embedding and kernel approximation, by optimizing the parameters of the projection matrix, we can significantly improve the performance. 

\subsection{The Need for Matrix $\mathbf{D}$}
\label{subsec:needD}
This matrix is required in prior work \cite{hinrichs2011johnson, vybiral2011variant, yu2014_cbe, yu2015compact}. By adding the random sign flipping matrix, the resulting projections are less correlated \cite{hinrichs2011johnson, vybiral2011variant}. In practice, the performance of a circulant neural network drops when the random sign flipping is not performed, which we demonstrate in Section \ref{subsec:withoutDexp}. 
To simplify the notation, we omit the matrix $\mathbf{D}$ in the following sections. 

\subsection{Space and Time Efficiency}
\label{subsec:space_time}
The two main advantages of the circulant binary embedding are superior space and time efficiency. 

\noindent\textbf{Space Efficiency.}
Typically, over 90\% of the storage cost of convolutional neural networks is due to the fully connected layers. So ``compressing'' such layers is a very important task. As the proposed circulant projection contains only $2d$ parameters ($d$ floats for $\bR$, and $d$ booleans for $\mathbf{D}$)\footnote{Note that the circulant matrix is never explicitly computed or stored.}, the space complexity is $\mathcal{O}(d)$.
This is a significant advantage compared to the conventional fully connected layer which requires $\mathcal{O}(d^2)$ parameters. For example, when $d = 4096$, as in the ``AlexNet'' \cite{krizhevsky2012}, the proposed method can decrease memory requirements by a factor of thousands, making the most space consuming component (the fully connected layer) negligible in memory cost.
We will show in Section \ref{sec:exp} that surprisingly, the circulant neural network can have very competitive performance  with such dramatic space savings. The small number of parameters also makes the neural network perform better with limited amount of training data. In addition, we can further improve the performance (yet maintain significant space savings) by adding more nodes (``fatter'' layers) and more layers. 

\noindent\textbf{Time Efficiency.}
The structure enables the use of Fast Fourier Transform (FFT) to speed up the computation. For $d$-dimensional data, the 1-layer circulant neural network has time complexity $\mathcal{O}(d \log d)$. Next we explain how we achieve this time complexity.
Given a data point $\mathbf{x}$, $h(\bx)$ can be efficiently computed as follows. 
Denote by $\circledast$ the operator of a circulant convolution. Based on the definition of a circulant matrix,
\begin{align}
\mathbf{R} \mathbf{x} = \br \circledast \mathbf{x}.
\end{align}
The convolution above can be computed more efficiently in the Fourier domain, using the Discrete Fourier Transform (DFT), for which a fast algorithm (FFT) is available. 
\begin{align}
h (\mathbf{x}) = \phi \left( \mathcal{F}^{-1} ( \mathcal{F}({\br}) \circ \mathcal{F}(\mathbf{x})) \right),
\end{align}
where $\mathcal{F}(\cdot)$ is the DFT operator, and $\mathcal{F}^{-1}(\cdot)$ is the inverse DFT (IDFT) operator.
As DFT and IDFT can be efficiently computed in $\mathcal{O}(d \log{d})$ with FFT \cite{oppenheim1999discrete}, the proposed approach has time complexity $\mathcal{O}(d \log{d})$. 

\subsection{When $k \ne d$}
\label{subsec:k}
We have so far assumed the number of nodes in the input layer $d$ to be equal to the number of nodes in the output layer.\footnote{Such a setting is commonly used in fully connected layers of recent convolutional neural network architectures.} 
In this section, we provide extensions to handle the case where $k \ne d$.

When $k<d$, the fully connected layer performs a ``compression'' of the signal. In this case, we still use the circulant matrix $\bR \in \mathbb{R}^{d \times d}$ with $d$ parameters, but the output is set to be the first $k$ elements in (\ref{eq:def}).  The circulant neural network is not computationally more efficient in this situation compared to when $k = d$.

When $k > d$, the fully connected layer performs an ``expansion'' of the signal. 
In this case, the simplest solution is to use multiple circulant projections, and concatenate their output. This has computational complexity $\mathcal{O}(k \log d)$, and space complexity $\mathcal{O}(k)$. Note that the DFT of the feature vector can be reused in this case. 
An alternative approach is to extend every feature vector to a $k$-dimensional vector, by appending $k-d$ zeros. This returns the problem to the previous setting described in section \ref{sec:framework} with $d$ replaced by $k$. This gives space complexity $\mathcal{O}(k)$, and computational complexity $\mathcal{O}(k \log k)$. 
In practice, $k$ is usually at most a few times larger than $d$. 
Empirically the two approaches take similar computational time. Our experimental results are based on the second approach.

\section{Training Circulant Neural Networks}
\label{sec:opt}
In this section, we propose a highly efficient way of training circulant neural networks. We also discuss a special type of circulant neural network, where the parameters of the circulant matrix are randomized instead of optimized. 

\subsection{Gradient Computation}
\label{subsec:grad}

The most critical step for optimizing a neural network given a training set is to compute the gradient of the error
function with respect to the network weights. Let us consider the conventional neural network with two layers,
where the first layer computes a linear transformation followed by a nonlinear activation function:
\begin{equation}
h(\bx) = \phi(\mathbf{R} \bx),
\end{equation}
where $\mathbf{R}$ is an unstructured matrix.
We assume the second layer is a linear classifier with weights $\mathbf{w}$.
Therefore the output of the two-layer neural network is
\begin{equation}
J(\bx) = \mathbf{w}^T \phi(\mathbf{R} \bx).
\label{eq:J}
\end{equation}

When training the neural network, computing the gradient of the error function involves computing the gradient of $J(\bx)$ with respect to each entry of $\mathbf{R}$. It is easy to show that
\begin{align}
&\frac{\partial J(\bx)}{\partial R_{ij}} = w_i \phi'(\bR_{i\cdot} \bx) x_j, \label{eq:gra}
\\
&i = 0, \cdots, d-1. \quad j = 0, \cdots, d-1.
\end{align}
where $\phi'(\cdot)$ is the derivative of $\phi(\cdot)$.

Note that (\ref{eq:gra}) suffices for the gradient-based optimization of neural networks, as the gradient \emph{w.r.t.} networks with more layers can simply be computed with the chain rule, leading to the well-known ``back-propagation'' scheme.

In the circulant case, we need to compute the gradient of the following objective function:
\begin{equation}
J(\bx) = \bw^T \phi(\bR \bx) = \sum_{i=0}^{d-1} w_i \phi \left( \mathbf{R}_{i\cdot} \bx \right),
\quad \mathbf{R} = \circR(\br).
\end{equation}
It is easy to show that
\begin{align}
\frac{\partial \bw^T \phi(\bR \bx)}{\partial r_i} 
&= \bw^T \left( \phi'(\bR \bx) \circ s_{\rightarrow i} (\bx) \right) \\
&= s_{\rightarrow i} (\bx)^T (\bw \circ \phi'(\bR \bx)),
\label{eq:gradient}
\end{align}
where $s_{\rightarrow i} (\cdot) : \mathbb{R}^d \rightarrow \mathbb{R}^d$, right (downwards for a column vector) circularly shifts the vector by one element. Therefore,
\begin{align}
&\nabla_{\br} J(\bx) \\
=& [s_{\rightarrow 0} (\bx), s_{\rightarrow 1} (\bx) 
, \cdots, s_{\rightarrow (d-1)} (\bx)]^T (\bw \circ \phi'(\bR \bx)) \nonumber\\
=& \circR(s_{\rightarrow 1} (\text{rev} (\bx) ))  (\bw \circ \phi'(\bR \bx)) \nonumber \\
=& s_{\rightarrow 1} (\text{rev} (\bx) ) \circledast (\bw \circ \phi'(\br \circledast \bx)), \nonumber
\end{align}
where,
\[
\text{rev} (\bx) = (x_{d-1}, x_{d-2}, \dots, x_0),
\]
\[
s_{\rightarrow 1} (\text{rev} (\bx) ) = 
(x_0, x_{d-1}, x_{d-2}, \cdots, x_1).
\]
The above uses the same trick of converting the circulant matrix multiplication to circulant convolution. 
Therefore, computing the gradient takes only $\mathcal{O}(d\log d)$ time with FFT. Training a multi-layer neural network is nothing more than applying (\ref{eq:gradient}) in each layer with the chain rule. 

Note that when $k < d$, we can simply set the last $d-k$ entries of $\bw$ in (\ref{eq:J}) to be zero. And when $k > d$, the above derivations can be applied with minimal changes. 

\subsection{Randomized Circulant Neural Networks}
\label{sec:rand}

We also consider the case where the elements of $\mathbf{r}$ in (\ref{eq:cir}) are generated independently from a standard normal distribution $\mathcal{N}(0,1)$. We refer
to these models as randomized circulant neural networks. In this case, the parameters of the circulant projections are defined by random weights, without optimization. In other words, in the optimization process, only the parameters of convolutional layers and the softmax classification layer are optimized.
This setting is interesting to study as it provides insight into the ``capacity'' of the model, independent of specific optimization mechanisms. 

We will show empirically that compared to unstructured randomized neural networks, the circulant neural network is faster with the same number of nodes, while achieving similar performance. This surprising result is in line with recent theoretical/empirical discoveries around using circulant projections on dimensionality reduction \cite{vybiral2011variant}, and binary embedding \cite{yu2014_cbe}.
It has been shown that the circulant projection behaves similarly to fully randomized projections in terms of the distance preserving properties.  In other words, the randomized circulant projection can be seen as a simulation of the unstructured randomized projection, both of which can capture global properties of the data.

In addition, we will show that with the optimizations described in Section \ref{subsec:grad}, the error rate of the neural networks improves significantly over the randomized version, meaning that the circulant structure is flexible and powerful enough to be used in a data-dependent fashion.

\section{Experiments}\label{sec:exp}
We apply our model to three standard datasets in our experiments: MNIST, CIFAR-10, and ImageNet. We note that it is not our goal to obtain state-of-the-art results on these datasets, but rather to provide a fair analysis of the effectiveness of circulant projections in the context of deep neural networks, compared to unstructured projections.

Next we describe our implementation and analysis of accuracy and storage costs on these three datasets, followed by an experiment on reduced training set size.

\begin{table*}
\begin{center}
\begin{tabular}{c|c|c|c|c}
\hline Method  &  Train Error & Test Error & Memory (MB) & Testing Time (sec.) \\ 
\hline LeNet   & 0.35\%  & 0.92\% & 1.56 & 3.06\\ 
\hline Circulant LeNet  & 0.47\% & 0.95\% & 0.27 & 2.14 \\ 
\hline 
\end{tabular}
\end{center}
\caption{Experimental results on MNIST.}
\label{tb:mnist}
\end{table*} 

\begin{table*}
\begin{center}
\begin{tabular}{c|c|c | c | c}
\hline Method  &  Train Error & Test Error & Memory (MB) & Testing Time (sec.) \\ 
\hline CIFAR-10 CNN  & 4.45\%  & 15.60\% & 0.45 & 4.56 \\ 
\hline Circulant CIFAR-10 CNN & 6.57\% & 16.71\%  & 0.12 & 3.92\\ 
\hline 
\end{tabular}
\end{center}
\caption{Experimental results on CIFAR-10.}
\label{tb:cifar}
\end{table*}

\subsection{Experiments on MNIST}
\label{subsec:mnist}

The MNIST digit dataset contains 60,000 training and 10,000 test images of ten handwritten digits (0 to 9), with $28 \times 28$ pixels. We use the LeNet network~\cite{Lecun98gradient-basedlearning} as our basic CNN model, which is known to work well on digit classification tasks. LeNet consists of a convolutional layer followed by a pooling layer, another convolution layer followed by a pooling layer, and then two fully connected layers similar to conventional multilayer perceptrons. We used a slightly different version from the original LeNet implementation, where the sigmoid activations are replaced by Rectified Linear Unit (ReLU) activations for the neurons. 

Our implementation extends Caffe \cite{jia2014caffe}, by replacing the weight matrix of the proposed circulant projections with the same dimensionality. The results are compared and shown in Table \ref{tb:mnist}. Our fast circulant neural network achieves an error rate of 0.95\% on the full MNIST test set, which is very competitive with the 0.92\% error rate from the conventional neural network. At the same time, the circulant LeNet is 5.7x more space efficient and 1.43x more time efficient than LeNet.

\subsection{Experiments on CIFAR}
\label{subsec:cifar}

CIFAR-10 is a dataset of natural 32x32 RGB images covering 10-classes with 50,000 images for training and 10,000 for testing. Images in CIFAR-10 vary significantly not only in object position and object scale within each class, but also in object colors and textures. 

The CIFAR10-CNN network \cite{dropout2012} used in our test consists of 3 convolutional layers, 1 fully-connected layer and 1 softmax layer. Rectified linear units (ReLU) are used as the activation
units. The circulant CIFAR10-CNN is implemented by using a circulant weight matrix to replace the fully connected layer. Images are cropped to 24x24 and augmented with horizontal flips, rotation, and scaling transformations. We use an initial learning rate of 0.001 and train for 700-300-50 epochs with their default weight decay. 

A comparison of the error rates obtained by circulant and unstructured projections is shown in Table \ref{tb:cifar}. Our efficient approach based on circulant networks obtains test error of 16.71\% on this dataset, compared to 15.60\% obtained by the conventional model. At the same time, the circulant network is 4x more space efficient and 1.2x more time efficient than the conventional CNN.

\begin{table*}
\begin{center}
\begin{tabular}{ c | c | c | c  }
\hline
Method                     & Top-5 Error   & Top-1 Error & Memory (MB) \\ \hline
Randomized AlexNet         & 33.5\%  &  61.7\%    & 233.2\\ \hline
Randomized Circulant CNN 1 & 35.2\%  &  62.8\%    & 20.5\\ \hline
AlexNet                    & 17.1 \% &  42.8\%    & 233.2\\ \hline
Circulant CNN 1            & 19.4 \% &  44.1\%    & 20.5\\ \hline
Circulant CNN 2            & 17.8 \% &  43.2\%    & 20.7\\ \hline
Reduced-AlexNet            & 37.2 \% &  65.3\%    & 20.7\\ \hline
\end{tabular}
\end{center}
\caption{Classification error rate and memory cost on ILSVRC-2010.}
\label{tb:large}
\end{table*}

\begin{table*}
\begin{center}
\begin{tabular}{l|c|c|c|c}
\hline $d$ & Full projection   & Circulant projection & Speedup  & Space Saving (in a fully connected layer)\\ 
\hline $2^{10}$ &   2.97  & 2.52 & 1.18x & 1,000x \\ 
\hline $2^{12}$ &   3.84  & 2.79 & 1.38x & 4,000x\\ 
\hline $2^{14}$ &   19.5  & 5.43 & 3.60x & 30,000x \\ 
\hline 
\end{tabular}
\end{center}
\caption{Comparison of training time (ms per image) and space of full projection and circulant projection. Speedup is defined as the time of circulant projection divided by the time of unstructured projection. Space saving is defined as the space of storing the circulant model by the space of storing the unstructured matrix. The unstructured projection matrix in conventional neural networks takes more than $90\%$ of the space cost. In AlexNet, $d$ is $2^{12}$.}
\label{table:training}
\end{table*}

\begin{figure*}
\centering 
\subfigure[MNIST]{\includegraphics[width=0.317\textwidth]{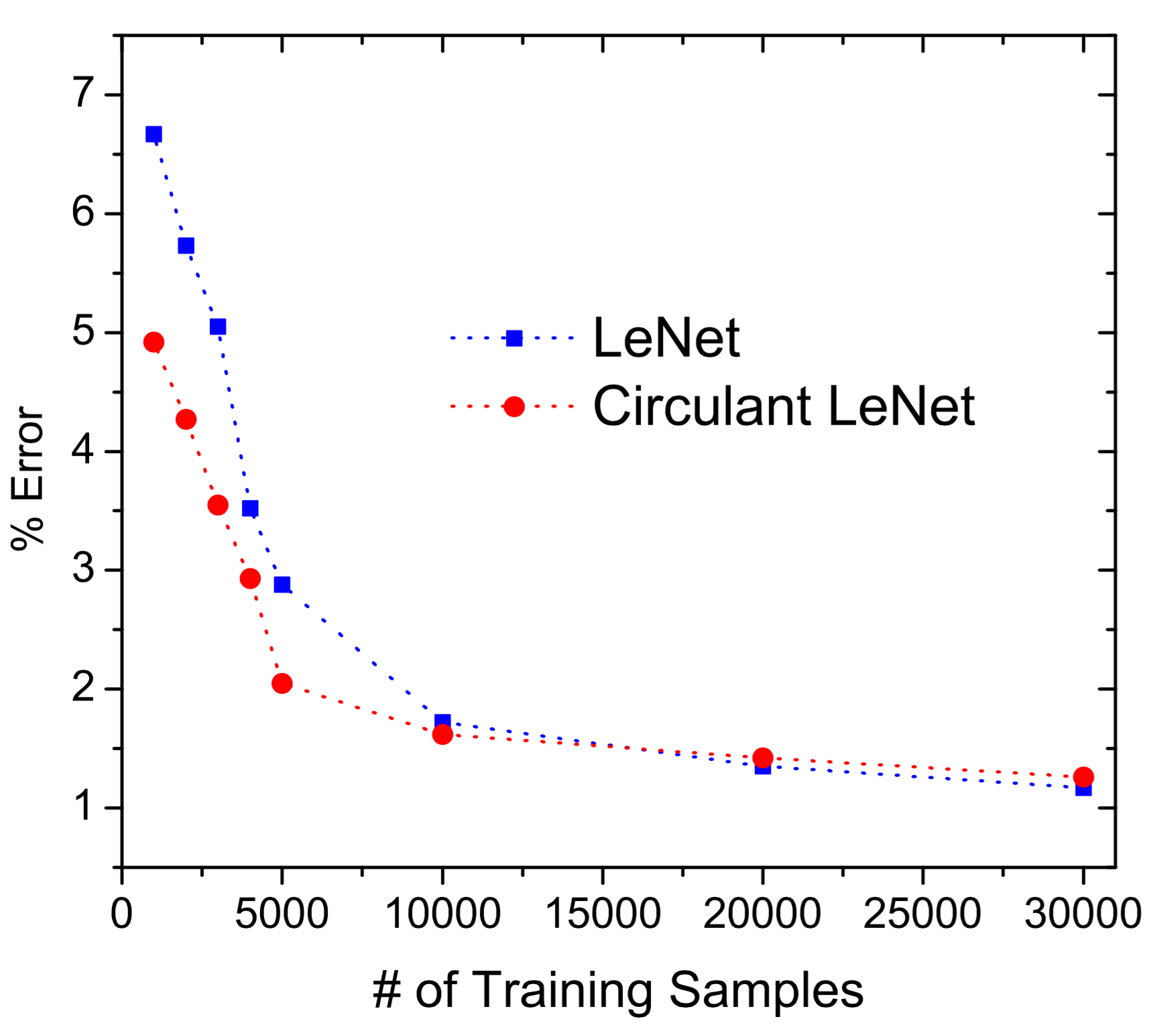}}
\subfigure[CIFAR-10]{\includegraphics[width=0.325\textwidth]{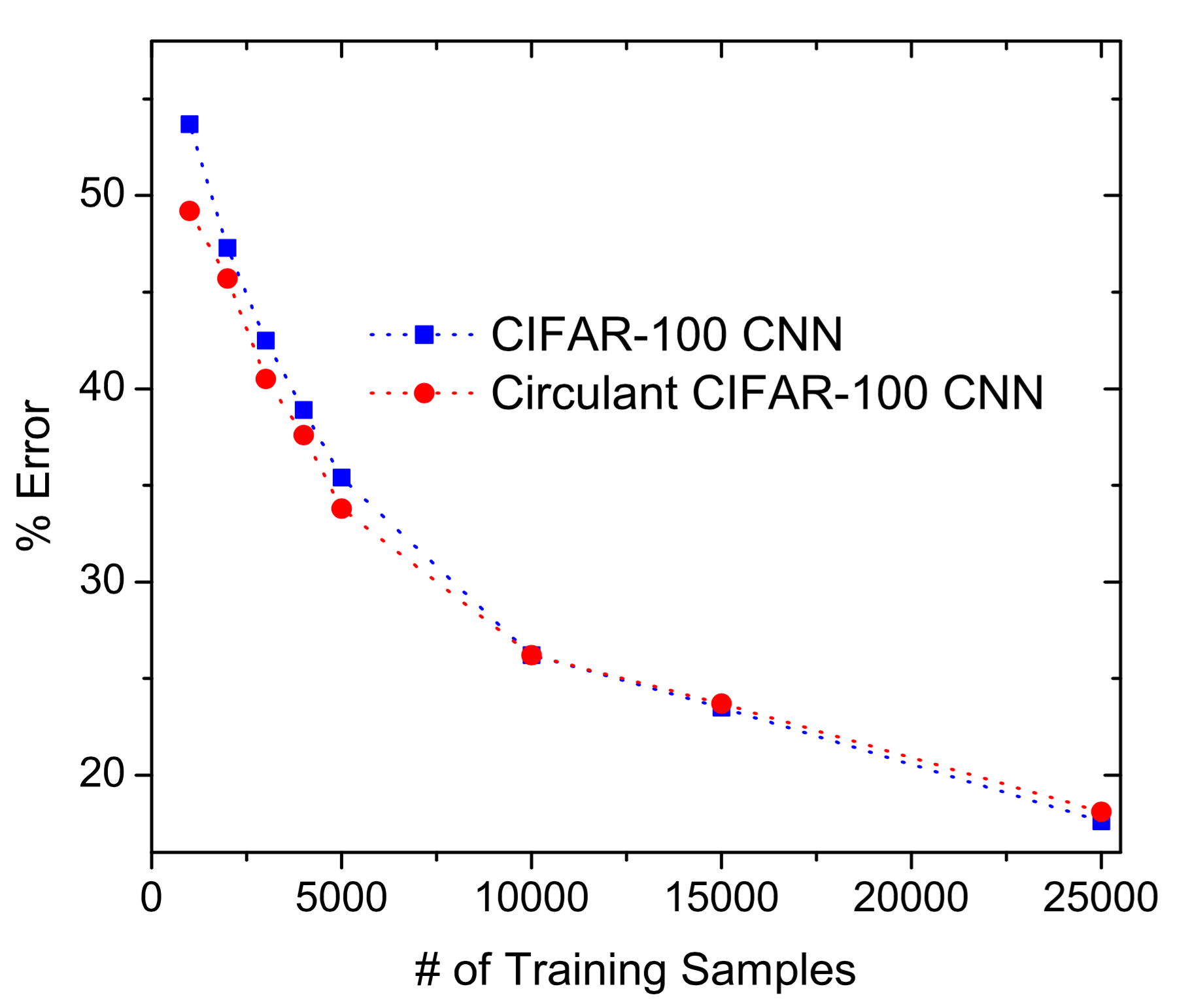}}
\subfigure[ILSVRC-2010]{\includegraphics[width=0.335\textwidth]{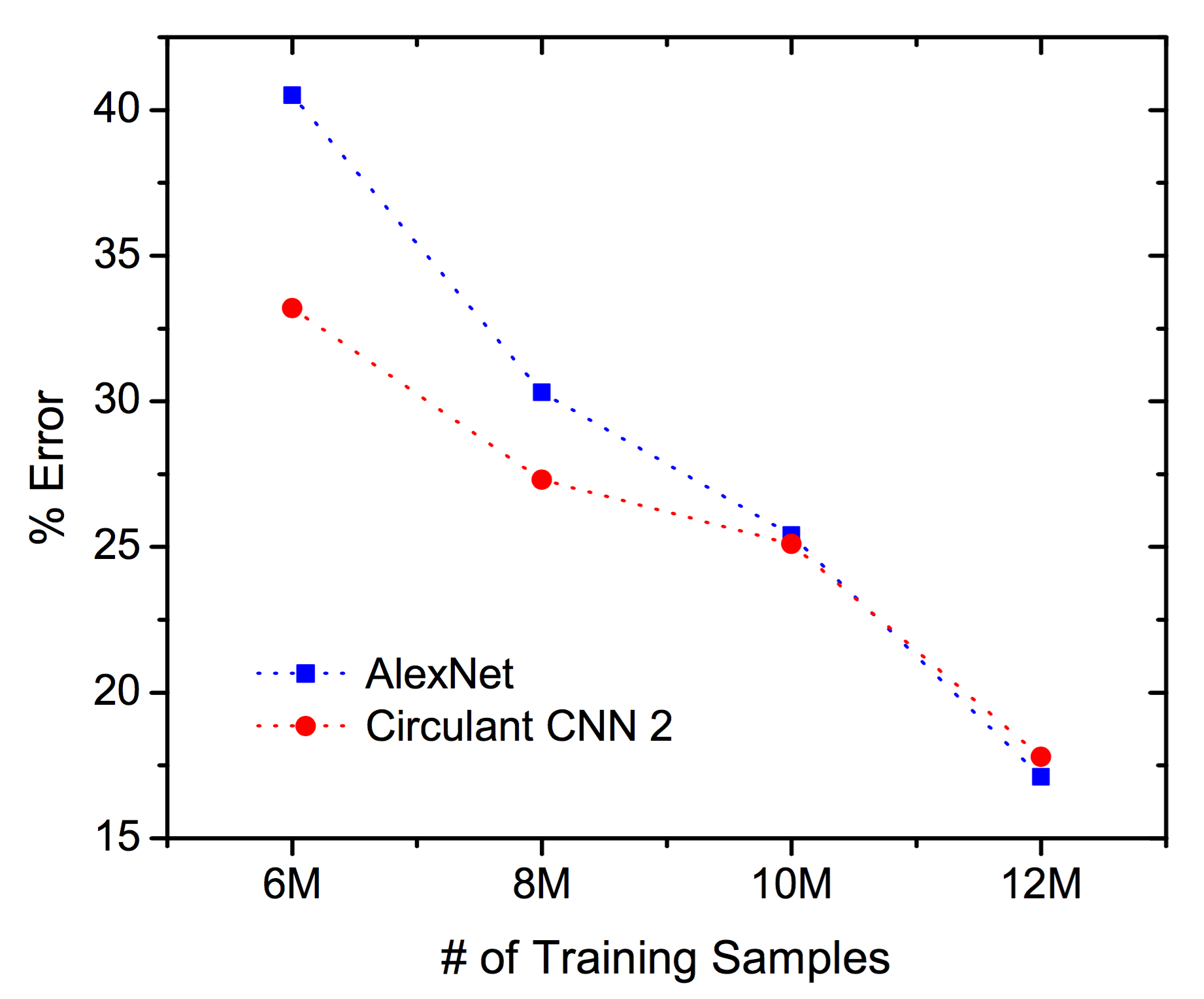}}
  \caption{Test error when training with reduced dataset sizes of circulant CNN and conventional CNN.}
\label{fig:sen}
\end{figure*}

\subsection{Experiments on ImageNet (ILSVRC-2010)}
\label{subsec:imagenet}
ImageNet is a dataset containing over 15 million labeled high-resolution images belonging to roughly 22,000 categories. 
Starting in 2010, as part of the Pascal Visual Object Challenge, an annual competition called the ImageNet Large-Scale Visual Recognition Challenge (ILSVRC) has been held. 
A subset of ImageNet with roughly 1000 images in each of 1000 considered categories is used in this challenge. Our experiments were performed on ILSVRC-2010.

We use a standard CNN network -- ``AlexNet'' \cite{krizhevsky2012} as the building block. The AlexNet consists of 5
convolutional layers, 2 fully-connected layers and 1 final softmax layer. Rectified linear units (ReLU) are used as the activation units. Pooling layers and response normalization layers are also used between convolutional layers. Our circulant network version involves three components: 1) a feature extractor, 2) fully circulant layers, and 3) a softmax classification layer. For 1 and 3 we utilize the Caffe package \cite{jia2014caffe}. For 2, we implement it with Cuda FFT.

All models are trained using mini-batch stochastic gradient descent (SGD) with momentum on batches of 128 images with the momentum parameter fixed at 0.9. 
We set the initial learning rate to 0.01, and manually decrease the learning rate if the network stops improving as in \cite{krizhevsky2012} according to a schedule determined on a validation set. Dataset augmentation is also exploited. 

Table \ref{tb:large} shows the error rate of various models.
We have used two types of structures for the proposed method. Circulant CNN 1 replaces the fully connected layers of AlexNet with circulant layers. 
Circulant CNN 2 uses ``fatter'' circulant layers compared to Circulant CNN 1: $d$ of Circulant CNN 2 is set to be $2^{14}$. 
In ``Reduced AlexNet'', we reduce the parameter size on the fully-connected layer of the original AlexNet to a size similar to our Circulant CNN by cutting $d$.
We have the following observations.
\begin{itemize}[leftmargin = *]
\item The performance of Randomized Circulant CNN 1\footnote{This is Circulant CNN 1 with randomized circulant projections. In other words, only the convolutional layer is optimized.} is very competitive with 
Randomized AlexNet. This is expected as the circulant projection closely simulates a fully randomized projection (Section \ref{sec:framework}). 
\item Optimization significantly improves the performance for both unstructured projections and circulant projections. The performance of Circulant CNN 1 is very competitive with AlexNet, yet with fraction of the space cost. 
\item By tweaking the structure to include more parameters, Circulant CNN 2 further drops the error rate to 17.8\%, yet it takes only a marginally larger amount of space compared to Circulant CNN 1, an 18x space saving compared to AlexNet. 
\item With the same memory cost, the Reduced AlexNet performs much worse than Circulant CNN 1.
\end{itemize}

In addition, one interesting finding is that ``dropout'', which is widely used in training CNNs, does not improve the performance of circulant neural networks. In fact it increases the error rate from 19.4\% (without dropout) to 20.3\% (not shown in the figure). This indicates that the proposed method is more immune to over-fitting.
We also show the training time (per image) on the standard and  circulant versions of AlexNet. We vary the number of hidden nodes $d$ in the fully connected layers and compare the training time until the model converges (ms/per image). Table \ref{table:training} shows the result. Our method provides dramatic space savings and significant speedup compared to the conventional approach.

\subsection{Reduced Training Set Size}
\label{subsec:size}
Compared to the neural network model with unstructured projections, the circulant neural network has fewer parameters. 
Intuitively, this may bring the benefit of better model generalization. In other words, the circulant neural network might be less data hungry compared to conventional neural networks. 
To verify our assumption, we report the performance of each model when trained with different training set sizes on the MINST, CIFAR-10, and ILSVRC-2010 datasets. Figure \ref{fig:sen} shows test error rate when training on a random subset of the training data. On MNIST and ILSVRC-2010, to achieve a fixed error rate, the circulant models need less data. On CIFAR-10, this improvement is limited as the circulant layer only occupies a small part of the model.

\subsection{Results Without $\mathbf{D}$}
\label{subsec:withoutDexp}
As noted in Section \ref{subsec:needD}, the sign flipping matrix $\mathbf{D}$ of (\ref{eq:def}) is important in our formulation. 
We provide some empirical results in this section. 
On the MNIST dataset, the test error rate increases from 0.95\% to 2.45\% by dropping $\mathbf{D}$. 
On the CIFAR-10 dataset, the test error rate increases from 16.71\% to 21.33\% by dropping $\mathbf{D}$.

\section{Discussion}
\subsection{Fully Connected Layer \vs Convolution Layer}
The goal of the method developed in this paper is to improve the efficiency of the fully connected layers of neural networks. 
In convolutional architectures, the fully connected layers are often the bottleneck in terms of the space cost. For example, in ``Alexnet'', the fully connected layers take $95\%$ of the storage. 
Remarkably, the proposed method enables dramatic space savings in the fully connected layer (4000x as shown in Table \ref{table:training}), making it  negligible in memory cost compared to the convolutional layers.
Our discovery resonates with recent work showing that the fully connected layers can be compressed, or even completely removed \cite{chen2014, szegedy2014}.

In addition, the fully connected layer costs roughly $20\%$ -- $30\%$ of the computation time based on our implementation. The FFT-based implementation can further improve the time cost, though not to the same degree as the realized space savings, if the majority of layers are convolutional. Our method is complementary to the work on improving the time and space cost of convolutional layers \cite{jaderberg2014,mathieu2013fast,erhan2014, denton2014exploiting}. 

\subsection{Circulant Projection \vs 2D Convolution}
One may notice that, although our approach leverages convolutions for speeding-up computations, it is fundamentally different from the convolutions performed in CNNs. The convolution filters in CNNs are all small 2D filters aiming at capturing local information of the images, whereas the proposed method is used to replace the fully connected layers, which are often ``big'' layers capturing global information. The operation involved is large 1D convolution rather than small 2D convolution.
The circulant projection can be understood as ``simulating'' an unstructured projection, with much less cost. Note that one can also apply FFT to compute the convolutions on the 2D convolutional layers, but due to the computational overhead, the speed improvement is generally limited on  small-scale problems. In contrast, our method can be used to dramatically speed up and scale the processing in fully connected layers. 
For instance, when the number of input nodes and output nodes are both $1$ million, conventional linear projection is essentially impossible, as it requires TBs of memory. On the other hand, doing a convolution of two $1$ million dimensional vectors requires only MBs of memory and tens of milliseconds.

\subsection{Towards Larger Neural Networks}
Currently, deep neural network models  usually contain hundreds of millions of  parameters. In real-world applications, there exist problems which involve an increasing amount of data.  We may need larger and deeper networks to learn better representations from large amounts of data. Compared to unstructured projections, the circulant projection significantly reduces computation and storage costs. Therefore, with the same amount of resources, circulant neural networks can use deeper as well as larger fully-connected networks. 
We have conducted preliminary experiments showing that the circulant model can be extended at least 10x deeper than conventional neural networks with the same scale of computational resources. 

\section{Conclusions}
We proposed to use circulant projections to replace the unstructured projections in order to optimize fully connected layers of neural networks. This dramatically improves the computational complexity from $\mathcal{O}(d^2)$ to $\mathcal{O}(d \log d)$ and space complexity from $\mathcal{O}(d^2)$ to $\mathcal{O}(d)$.
An efficient approach was proposed for optimizing the parameters of the circulant projections. 
We demonstrated empirically that this optimization can lead to much faster convergence and training time compared to conventional neural networks. 
Our experimental analysis was carried out on three standard datasets, showing the effectiveness of the 
proposed approach. We also reported experiments on randomized circulant projections, achieving performance similar to that of unstructured randomized projections. Our ongoing work is to explore different matrix structures to compress and speed up neural networks.

\noindent\textbf{Acknowledgement.} Felix X. Yu was supported in part by the IBM PhD Fellowship Award when working on this project at Columbia University. We would like to thank Dan Holtmann-Rice for proofreading the paper.

{\small
\bibliographystyle{ieee}
\bibliography{design}
}

\end{document}